# Pruning Distorted Images in MNIST Handwritten Digits

Amarnath R
EdTech Businesses, HCL Technologies Ltd, Elcot Sez, Sholinganallur, Chennai, India.
amarnathresearch@gmail.com
and
Vinay Kumar V
IVIS Labs Pvt. Ltd, Harju maakond, Tallinn, Kesklinna linnaosa, Ahtri tn 12, 15551, Estonia.
vinay@ivislabs.com

## ABSTRACT

Recognizing handwritten digits is a challenging task primarily due to the diversity of writing styles and the presence of noisy images. The widely used MNIST dataset, which is commonly employed as a benchmark for this task, includes distorted digits with irregular shapes, incomplete strokes, and varying skew in both the training and testing datasets. Consequently, these factors contribute to reduced accuracy in digit recognition. To overcome this challenge, we propose a two-stage deep learning approach. In the first stage, we create a simple neural network to identify distorted digits within the training set. This model serves to detect and filter out such distorted and ambiguous images. In the second stage, we exclude these identified images from the training dataset and proceed to retrain the model using the filtered dataset. This process aims to improve the classification accuracy and confidence levels while mitigating issues of underfitting and overfitting. Our experimental results demonstrate the effectiveness of the proposed approach, achieving an accuracy rate of over 99.5% on the testing dataset. This significant improvement showcases the potential of our method in enhancing digit classification accuracy. In our future work, we intend to explore the scalability of this approach and investigate techniques to further enhance accuracy by reducing the size of the training data.

## 1 INTRODUCTION

Handwritten digit recognition is a complex task that finds applications in various fields, including computer vision and machine learning. It involves the identification and classification of digits written by hand, enabling tasks such as character recognition and digit analysis. In this domain, the MNIST dataset serves as a widely used benchmark for evaluating the performance of handwritten digit recognition systems [1, 2]. It consists of a collection of 70,000 handwritten digits, ranging from 0 to 9, and is derived from the writings of 250 individuals. The dataset's standardized characteristics, such as the consistent image size of 28x28 pixels and grayscale format, make it particularly valuable for research and experimentation [1, 2, 5]. While there are other handwritten datasets available in the literature [3], the MNIST dataset stands out due to its comprehensive nature. It provides a diverse range of writing styles, capturing variations in how different individuals write digits. Moreover, the dataset contains samples with various distortions, representing real-world scenarios where digits can appear in different shapes or orientations. This diversity in the dataset enables researchers to explore and develop robust algorithms that can handle these challenges effectively.

Although deep learning techniques have achieved significant success in digit classification on the MNIST dataset, achieving maximum accuracy remains a challenging task [4]. This challenge arises from various factors that can impact the model's ability to learn and classify digits accurately. One of these factors is data imbalance, where certain classes of digits may have a disproportionately small number of samples compared to others. This imbalance can hinder the model's performance as it may struggle to learn effectively from limited data, leading to biased predictions. In addition to data imbalance, variances in the data itself can also present challenges. Handwritten digits can exhibit irregular shapes, incomplete strokes, and variations in their height, width, density, concavity, and convexity. These variations make it difficult for the model to generalize and capture the essential features necessary for accurate classification. However, the most significant obstacle faced by researchers is the identification of ambiguous images within the training set. These ambiguous images can introduce noise during the training process, leading to issues of overfitting or underfitting and ultimately compromising the accuracy of the model [4, 5]. When ambiguous images are present in the training set, the model may learn to rely on specific artifacts or noise patterns rather than the true characteristics of the digits. As a result, the model becomes biased and lacks the ability to generalize well for the specific task at hand.

Data preparation plays a crucial role in building an effective deep network pipeline. However, manually removing noisy images, particularly in large training sets, can be challenging and time-consuming [6, 7]. Figure 1 showcases a selection of ambiguous images randomly extracted from the MNIST training dataset, which includes digits ranging from 0 to 9. The figure displays five sets of randomly chosen samples, organized in columns, where the first column represents the digit 0, and the last column represents the digit 9. Upon observing the figure, it becomes evident that a significant portion of the images is inherently ambiguous, making accurate classification a difficult task even for human observers. These ambiguous images present challenges due to their unclear or distorted visual characteristics. The presence of irregular shapes, incomplete strokes, variations in size, or conflicting elements within the images can cause confusion and lead to difficulties in accurate classification. These challenges highlight the need for automated approaches to identify and remove such ambiguous images from the training dataset, as manual inspection and removal would be labor-intensive and impractical, especially in scenarios involving large datasets.

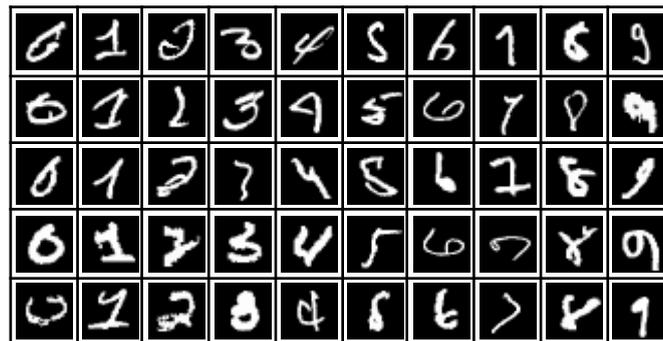

Figure 1: Five sets of samples consisting of digits ranging from 0 to 9 from the training dataset.

The objective of our research is to detect and eliminate distorted images that cause ambiguity within the training set. However, manually identifying such images can be a challenging task. To overcome this challenge, we propose the use of a deep learning model initially. This model is trained to identify and flag these distorted images automatically. Once the distorted images are identified, we proceed to verify them through human inspection. This step ensures that the flagged images are indeed ambiguous and warrant removal from the training set. After the verification process, the model is retrained using the filtered images from the training set, which no longer include the identified distorted images. In this paper, we present a deep neural network architecture specifically designed to identify and remove such distorted images from the MNIST dataset.

Our approach aims to improve the overall accuracy of the digit classification model by reducing variations caused by ambiguous images. By filtering out these images, we enhance the quality of the training data, leading to more reliable and accurate predictions. Furthermore, our approach increases confidence levels in the training and validation losses, providing a more robust assessment of the model's performance. It also addresses issues related to overfitting and underfitting, which can occur when the model is excessively influenced by the presence of distorted or ambiguous images during training. The inspiration for our approach comes from previous studies [6, 7, 8] that have tackled similar challenges in data preparation and noise removal. By building upon the knowledge and techniques established in these studies, we aim to contribute to the field of digit classification and further improve the accuracy and reliability of deep learning models.

Previous research [2, 9] has shown that deep learning techniques can significantly improve handwritten digit recognition accuracy. For instance, LeCun et al. [2] achieved an error rate of 0.7% using convolutional neural networks (CNNs) for digit recognition. Later studies reported even better results using more advanced deep learning architectures, such as ResNet (He et al., 2016) [10] and Capsule Networks (Sabour et al., 2017) [11]. However, these works have not extensively addressed the issue of identifying and eliminating such distorted images causing lower accuracy rates.

In recent studies, researchers have introduced various approaches [12, 13, 14, 15] aimed at identifying and removing noisy, ambiguous, or distorted images from datasets used in different image classification tasks. These techniques have demonstrated success in enhancing the quality of training data and improving classification accuracy. Motivated by the positive outcomes of these techniques, our research aims to apply similar methods to the MNIST dataset. Specifically, we focus on detecting and eliminating ambiguous images that can introduce noise during the training process. By addressing this issue, we aim to enhance the reliability and performance of the classification model. To achieve our goal, we propose a deep neural network architecture that encompasses preprocessing, feature extraction, and classification layers. By following a systematic approach, our method aims to improve classification accuracy, ensure an unbiased and generalized model, and enhance confidence levels in the model's predictions. The key steps involved in our approach are as follows:

1. Development of a model capable of detecting ambiguous images in the training dataset. This model serves as a means to automatically identify and flag images that exhibit characteristics of ambiguity.
2. Manual validation and removal of the identified ambiguous images from the training dataset. By conducting a thorough inspection, we ensure the accurate identification of these images and their subsequent exclusion from the training data.
3. Retraining the deep neural network using the remaining images from the training dataset. This step allows the model to learn from a refined dataset, free from the noise introduced by the ambiguous images. By retraining the model, we aim to improve its classification accuracy and boost confidence levels in its predictions.
4. Reporting our findings, including accuracies and losses, after cleaning the training dataset. We present the outcomes of our approach, showcasing the impact of removing ambiguous images on the model's performance and highlighting the improvements achieved.

Our approach offers the following significant contributions:

- Improved Confidence Level: Our approach has successfully enhanced the confidence level of predictions for digit recognition. By detecting and removing ambiguous images, which are challenging even for humans to recognize, we have increased the reliability and trustworthiness of our model's predictions.
- Increased Accuracy: Through the identification and elimination of ambiguous images, we have achieved a higher accuracy rate in our model. By removing these sources of confusion and noise from the training dataset, our model can focus on learning from more reliable and representative examples, leading to improved classification performance.
- Effective Neural Network Design: Our neural network design is straightforward yet effective in addressing both underfitting and overfitting issues. By carefully designing the architecture and training process, we have created a model that achieves a balance between capturing important patterns in the data while avoiding overgeneralization or memorization.
- High Accuracy without Augmented Data: Remarkably, we have achieved high accuracy on the MNIST dataset without utilizing any augmented data for training. Our model's performance demonstrates its capability to learn and generalize from the available dataset effectively. Moreover, we have also demonstrated that our model performs well even with a reduced dataset, suggesting its robustness and adaptability.

The paper is structured as follows: Section 2 provides a comprehensive overview of the existing literature on handwritten digit classification in the MNIST dataset. Section 3 describes the MNIST dataset's characteristics, including the data distribution, image characteristics, and classification challenges. In Section 4, we present our proposed method in detail, including the deep neural network architecture, preprocessing steps, and the identification and removal of ambiguous images. Section 5 presents experimental results. Finally, in Section 6, we summarize the paper's contributions and conclude our study.

## 2 RELATED WORKS

In this survey, we will review the most significant studies related to and limited to the MNIST digit classification using deep learning.

Convolutional neural networks have been widely used for MNIST digit classification, achieving high accuracy rates. LeNet-5, introduced by Yann LeCun et al. [2] in 1998, was the first successful CNN for handwritten digit recognition. LeNet-5 includes two convolutional layers, two subsampling layers, and three fully connected layers. Another popular CNN is AlexNet, which won the ImageNet Large Scale Visual Recognition Challenge (ILSVRC) in 2012 [16]. AlexNet includes five convolutional layers, three fully connected layers, and a ReLU activation function. Other successful CNN models include VGG-Net[17], GoogLeNet[18], ResNet[10], and InceptionNet [20].

In 2019, Baldominos et al. [21] conducted a literature survey on the research and developments related to the MNIST dataset. In our study, we have examined the surveys mentioned in their paper that report a test error rate below 0.5. Specifically, we present the findings in Table 1, which corresponds to test accuracies achieved with augmentation, and Table 2, which represents test accuracies without augmentation. Their contributions have been discussed in the literature [21]

Table 1: Side-by-side comparison of the most competitive (error rate < 0.5%) results found in the state of the art for the MNIST database with data augmentation or preprocessing

| Technique | Test Datset Error Rate < 0.5% |
|---|---|
| NN 6-layer 5,700 hidden units[22] | 0.35% |
| MSRV C-SVDDNet [23] | 0.35% |
| Committee of 25 NN 2-layer 800 hidden units [24] | 0.39% |
| RNN [25] | 0.45% |
| CNN (2 conv, 1 dense, relu) with DropConnect [26] | 0.21% |
| Committee of 25 CNNs [27] | 0.23% |
| CNN with APAC [28] | 0.23% |
| CNN (2 conv, 1 relu, relu) with dropout [26] | 0.27% |
| Committee of 7 CNNs [30] | 0.27% |
| Deep CNN [31] | 0.35% |
| CNN (2 conv, 1 dense), unsup pretraining [32] | 0.39% |
| CNN, XE loss [33] | 0.40% |
| Scattering convolution networks + SVM [34] | 0.43% |

Table 2: Side-by-side comparison of the most competitive (error rate < 0.5%) results found in the state of the art for the MNIST database without data augmentation or preprocessing

| Technique | Test Dataset Error Rate < 0.5% |
|---|---|
| Batch-normalized maxout network-in-network [35] | 0.24% |
| Committees of evolved CNNs (CEA-CNN) [36] | 0.24% |
| Genetically evolved committee of CNNs [37] | 0.25% |
| Committees of 7 neuroevolved CNNs [38] | 0.28% |
| CNN with gated pooling function [39] | 0.29% |
| Inception-Recurrent CNN + LSUV + EVE [40] | 0.29% |
| Recurrent CNN [41] | 0.31% |
| CNN with norm. layers and piecewise linear activation units [42] | 0.31% |
| CNN (5 conv, 3 dense) with full training [43] | 0.32% |

In a recent study published in 2022, researchers [44] utilized a decoder network, which proved to be a more convenient and quicker training approach when compared to encoder-decoder architectures. However, the authors did not mention the presence of noisy images in the MNIST dataset.

In contrast, a more recent study published in 2023 by authors [45] explicitly addressed the issue of distorted images by incorporating samples from the MNIST dataset. They demonstrated how these distorted images can significantly decrease the accuracy rate, even when using state-of-the-art deep learning networks. The researchers emphasized the limitations of current deep learning models in comparison to human capabilities, highlighting that existing image distortions used for evaluating deep learning models primarily rely on mathematical transformations rather than human cognitive functions.

Furthermore, the 1998 paper written by Yann LeCun et al. [2] brings attention to the existence of genuinely ambiguous and underrepresented digits within the training set, which presents difficulties in their recognition.

Based on the literature survey mentioned earlier, it has been noted that only a limited number of papers discuss distorted images within the MNIST dataset. However, there is a lack of comprehensive research dedicated to identifying and removing such images from the training set. Inspired by the recent work highlighted in papers [45, 46] and foundation paper [2], we cognize the implications of these noisy images on compromising accuracy. Consequently, in this study, we propose a deep neural network architecture explicitly designed to eliminate ambiguous images from the MNIST training dataset. Our approach effectively addresses issues of overfitting and underfitting by minimizing variations in training and validation losses, resulting in enhanced accuracy for the digit classification model. By tackling the crucial task of identifying and removing ambiguous images, our methodology provides a practical solution that elevates the overall accuracy of deep learning models utilized for handwritten digit recognition.

## 3 MNIST DATASET

The MNIST dataset is a widely used benchmark dataset in the field of image classification. It was introduced in a research paper by Yann LeCun et al [1, 2]. The dataset consists of 70,000 images of handwritten digits, which are split into 60,000 training images and 10,000 test images. Each image is a grayscale 28x28 pixel image, centered on a white background. The digits are written by 250 individuals, which results in variations in writing styles, orientations, and distortions. The goal of the MNIST digit classification problem is to develop an algorithm that can accurately classify the digits in the images into their corresponding classes (0-9). The MNIST dataset is considered a challenging problem due to the variability in the dataset like disconnected or incomplete strokes, skewed or rotated digits with irregular shapes, which makes it difficult to accurately classify the digits.

The MNIST dataset, although widely popular, is not immune to challenges. One significant obstacle is the existence of ambiguous images that can affect classification model accuracy [45]. Such images are often difficult for both humans and algorithms to distinguish between certain digits [46], such as 4 and 9, 1 and 7, 6 and 0, 7 and 4, and so on. When such noisy or distorted images are included in the training dataset, the learning rate can decrease, and the model's accuracy can be compromised. Figure 2 illustrates some of these images, where the digits from 0 to 9 are displayed in ascending order.

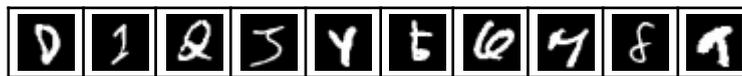
Figure 2: A sample set of noisy images from training data

In addition to the presence of noisy images, the MNIST dataset poses other challenges as well. One such challenge is the variations in writing styles and quality among different individuals. This results in some digits being more difficult to recognize than others, making the dataset more complex for classification models. Additionally, the dataset is imbalanced, with certain digits having significantly more samples than others [2, 47, 50, 51]. This can create bias in the model towards the more frequently occurring digits and poorer classification performance on less common digits. The distribution of samples across the 10 classes is illustrated in Figure 3.

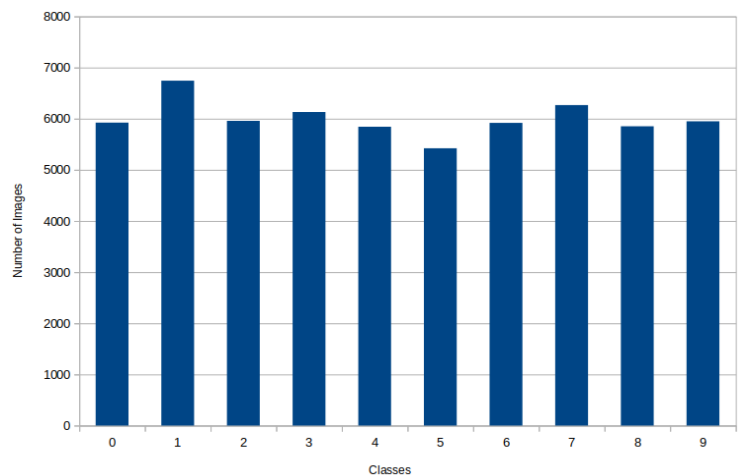
Figure 3: Distribution of digits in Training Set

## 4 PROPOSED NETWORK DESIGN

The proposed method consists of three components: a preprocessing layer, a feature extraction layer, and a classification layer, as depicted in Figure 4. We used the TensorFlow and Keras frameworks for this study, as they are both widely used and make it easy to develop deep learning models [48, 49].

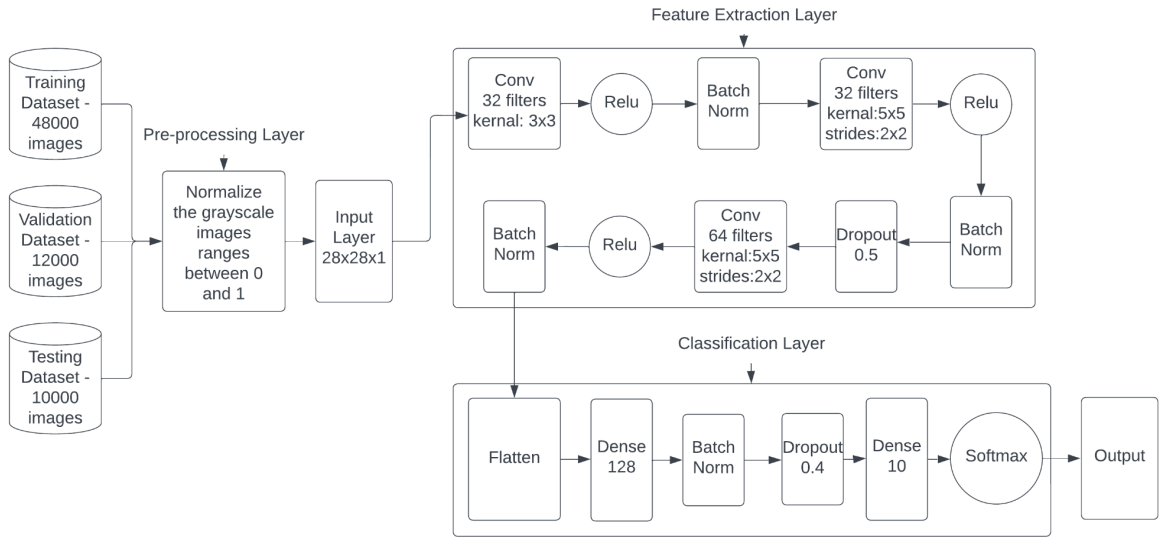

Figure 4: Deep Learning Framework

Figure 4 shows that the preprocessing layer performs normalization of the input images to a standard size of 28x28 and scales the pixel values between 0 and 1. This step helps to reduce the variations in input images, making them more suitable for training deep neural networks.

The feature extraction layer includes 3 convolutional layers, each with the ReLu activation function and Batch Normalization applied to the output. To address overfitting, we also used the Dropout layer in combination with Batch Normalization. Dropout regularization randomly sets a fraction of the neurons to zero during training, which helps to prevent over-reliance on specific features. Batch normalization normalizes the activations of the neurons to reduce the impact of internal covariate shifts, which can improve generalization performance.

The classification layer is composed of multiple connected layers that map the extracted features to the output classes. To tackle the overfitting problem, we introduce dropout regularization and batch normalization in the fully connected layers. Dropout regularization randomly sets a fraction of the neurons to zero during training, while batch normalization normalizes the activations of the neurons to reduce the internal covariate shift. Figure 5 displays the layers and the learnable parameters of the proposed architecture.

```
Layer (type)                    Output Shape              Param #
=================================================================
convolution l (InputLayer)      [(None, 28, 28, 1)]       0

conv2d (Conv2D)                 (None, 26, 26, 32)        320

batch_normalization (BatchN     (None, 26, 26, 32)        128
ormalization)

conv2d_1 (Conv2D)               (None, 13, 13, 32)        25632

batch_normalization_1 (Batc     (None, 13, 13, 32)        128
hNormalization)

dropout (Dropout)               (None, 13, 13, 32)        0

conv2d_2 (Conv2D)               (None, 7, 7, 64)          51264

batch_normalization_2 (Batc     (None, 7, 7, 64)          256
hNormalization)

dropout_1 (Dropout)             (None, 7, 7, 64)          0

flatten (Flatten)               (None, 3136)              0

dense (Dense)                   (None, 128)               401536

batch_normalization_3 (Batc     (None, 128)               512
hNormalization)

dropout_2 (Dropout)             (None, 128)               0

dense_1 (Dense)                 (None, 10)                1290

=================================================================
Total params: 481,066
Trainable params: 480,554
Non-trainable params: 512
```

Figure 5: Layers of deep network with parameters

Figure 5 displays the layers and learnable parameters of the proposed architecture. We can see that the feature extraction component consists of multiple layers that extract relevant features from the input grayscale images of dimensions 28x28. The first layer serves as the input layer and has no learnable parameters, making it a non-trainable layer. The second layer is a convolutional layer with 32 filters, a kernel size of 3x3, no padding, and a stride of 1x1. The number of trainable parameters in this layer is 320. The third layer is a batch normalization layer with 32 filters and 128 trainable parameters, which include gamma weights, beta weights, moving_mean, and moving_variance. The fourth layer is another convolutional layer with 32 filters, a kernel size of 5x5, no padding, a stride of 2x2, and 25,632 trainable parameters. The fifth layer is another batch normalization layer with 128 trainable parameters. The sixth layer is a dropout layer with a dropout rate of 0.5 and no trainable parameters. The seventh layer is a convolutional layer with 64 filters, a kernel size of 5x5, no padding, a stride of 2x2, and 51,264 trainable parameters. The eighth layer is another batch normalization layer with 256 trainable parameters. The ninth layer is another dropout layer with a dropout rate of 0.5 and no trainable parameters. The final layer in the feature extraction component is a flattened layer with 3,136 feature maps and no trainable parameters.

Following the feature extraction component, the classification component of the neural network consists of four layers. The first layer is a dense layer with 128 neurons and 401,536 trainable parameters. The second layer is a batch normalization layer with 512 trainable parameters. The third layer is a dropout layer with a value of 0.4 and no trainable parameters. The final layer is a dense layer with the softmax activation function and 1,290 trainable parameters.

Here are the trainable and non-trainable parameters for each layer along with the formula to compute them:

Feature extraction layer:
- Input layer: non-trainable (no learnable parameters)
- Conv2D layer: 320 trainable parameters.
  Formula: $(w * h + 1) * f = (3 * 3 + 1) * 32 = 320$
- BatchNormalization layer: 128 trainable parameters.
  Formula: $f * 4 = 32 * 4 = 128$
- Conv2D layer: 25,632 trainable parameters.
  Formula: $(w * h * f_p + 1) * f = (5 * 5 * 32 + 1) * 32 = 25,632$
- BatchNormalization layer: 128 trainable parameters.
  Formula: $f * 4 = 32 * 4 = 128$
- Dropout layer: non-trainable (no learnable parameters)
- Conv2D layer: 51,264 trainable parameters.
  Formula: $(w * h * f_p + 1) * f = (5 * 5 * 32 + 1) * 64 = 51,264$
- BatchNormalization layer: 256 trainable parameters.
  Formula: $f * 4 = 64 * 4 = 256$
- Dropout layer: non-trainable (no learnable parameters)
- Flatten layer: non-trainable (no learnable parameters)

Classification layer:
- Dense layer: 401,536 trainable parameters.
  Formula: $(n_p * n) + (1 * n) = 3136 * 128 + 128 = 401,536$
- BatchNormalization layer: 512 trainable parameters.
  Formula: $f * 4 = 128 * 4 = 512$
- Dropout layer: non-trainable (no learnable parameters)
- Dense layer: 1,290 trainable parameters.
  Formula: $(n_p * n) + (1 * n) = 128 * 10 + 10 = 1,290$

In the above formulas, "$w$" and "$h$" represent the width and height of the kernel, "$f$" represents the number of filters in the current layer, "$f_p$" represents the number of filters in the previous layer, and "$n$" and "$n_p$" represent the number of neurons in the current and previous layers, respectively. The "$+ 1$" term in the formulas represents the bias term.

Note that the batch normalization layers have both trainable and non-trainable parameters. The trainable parameters include "gamma" and "beta" weights used to adjust the normalized data, while the non-trainable parameters include "moving_mean" and "moving_variance" used to keep track of the mean and variance of the normalized data across all mini-batches during training.

In summary, the proposed neural network architecture has a total of 480,554 trainable parameters and 512 non-trainable parameters.

The neural network consists of 3 convolution layers and 2 dense layers with batch normalization and dropouts. This layer setup allows the system to learn more features and handle both overfitting and underfitting issues effectively. However, adding extra convolution layers or filters can lead to overfitting, while reducing the number of filters or convolution layers can result in underfitting. Therefore, we have maintained this layer setup to ensure that the network learns sufficient features.

To further handle overfitting, we have also incorporated significant dropout, which helps match the validation losses with the training losses. This choice of layer and the number of trainable parameters provides a clear understanding of the neural network's design. Each layer is discussed in detail below.

In our neural network, convolutional layers were employed to identify the spatial features of the images, while batch normalization layers were used to stabilize the learning process and improve training time. To achieve this, we chose kernel sizes of 3x3 and 5x5, which are effective for filtering corner edges and lengthy lines, respectively. Additionally, we selected filter sizes 32 and 64 to enable more learning parameters, which can lead to overfitting if not handled properly.

To prevent overfitting and enhance the generalization ability of the model, we added dropout layers. These layers randomly drop out some nodes during training, preventing the model from relying too much on specific features or patterns in the data. This further improves the network's ability to handle noisy or ambiguous images and obtain accurate results.

To compute the trainable parameters for each layer, we derived a formula based on the number of input channels, output channels, kernel size, and other hyperparameters used in that particular layer. The non-trainable parameters, such as the moving_mean and moving_variance in the batch normalization layers, were used to track the mean and variance of the input data during training, which is essential for the normalization process.

The proposed neural network architecture and the number of trainable and non-trainable parameters used were determined based on previous research and experimentation in the field of image classification. The deep network architecture with multiple layers and a large number of trainable parameters enables the model to learn complex features and patterns in the input images, which can improve the accuracy of the classification task.

Overall, our proposed neural network architecture provides a novel approach to image classification that considers the specific spatial and structural features of the images. Through the careful selection of layers and hyperparameters, we demonstrate the potential for improving the accuracy of image classification tasks.

Determining optimal hyperparameters for a neural network is an important step in achieving high accuracy on a given dataset. Below are hyperparameters used for the proposed neural network architecture for classifying the MNIST dataset:
1. Learning rate: A good starting point for the learning rate is 0.0001.
2. Batch size: The batch size determines the number of samples that are processed at once during training. As we have used batch normalization, we recommended using a batch size of 128 which provides better accuracy.
3. Number of epochs: As the MNIST dataset is 60000 images, we have experimented with having 200 epochs. However, we find that early stops at ~90 epochs to get better accuracy having patients at 7.
4. Dropout rate: We have dropouts which are 0.4 and 0.5 at different positions just to address the overfitting problem.
5. Number of filters: A good starting point for the number of filters is 32 or 64. However, the optimal number of filters may depend on the complexity of the dataset and the size of the network. Accordingly, the filters are used.

It's worth noting that determining optimal hyperparameters can be a time-consuming and iterative process, and it may be necessary to try different combinations of hyperparameters to find the best values for a given dataset.

## 5  EXPERIMENTAL RESULTS

As mentioned in Section 2, the MNIST dataset consists of 70,000 grayscale images of handwritten digits from 0 to 9, with 60,000 images reserved for training and 10,000 images for testing. The images are of size 28x28 pixels, and each pixel is represented by a single value between 0 and 255, indicating the grayscale intensity.

## 5.1 STAGE 1: TRAINING & TESTING WITH COMPLETE DATA

To improve the training process, the 60,000 training images were further divided into two subsets: 80% (48,000 images) were used for actual training, and 20% (12,000 images) were allocated for validation purposes. This division allowed for monitoring the model's performance on an unseen dataset and adjusting the model's hyperparameters accordingly. The choice of an 8:2 ratio split is consistent with prior research papers [52, 53], although this ratio can be altered, and investigating alternative ratios falls outside the scope of this paper and opens an area for future research.

To ensure that the testing images were not utilized during training or validation, they were excluded from both the training and validation sets. K-fold cross-validation with n_splits=5 parameter was used to split the training and validation data five times without shuffling the data. This approach ensured that there were no duplicates in the split sets, and the model could be evaluated thoroughly.

The training, validation accuracy, and loss graphs were plotted to assess the model's performance, as shown in Figure 6. The variance between the training and validation losses and accuracy was minimal from epoch 40, indicating that the model did not have any issues with overfitting or underfitting.

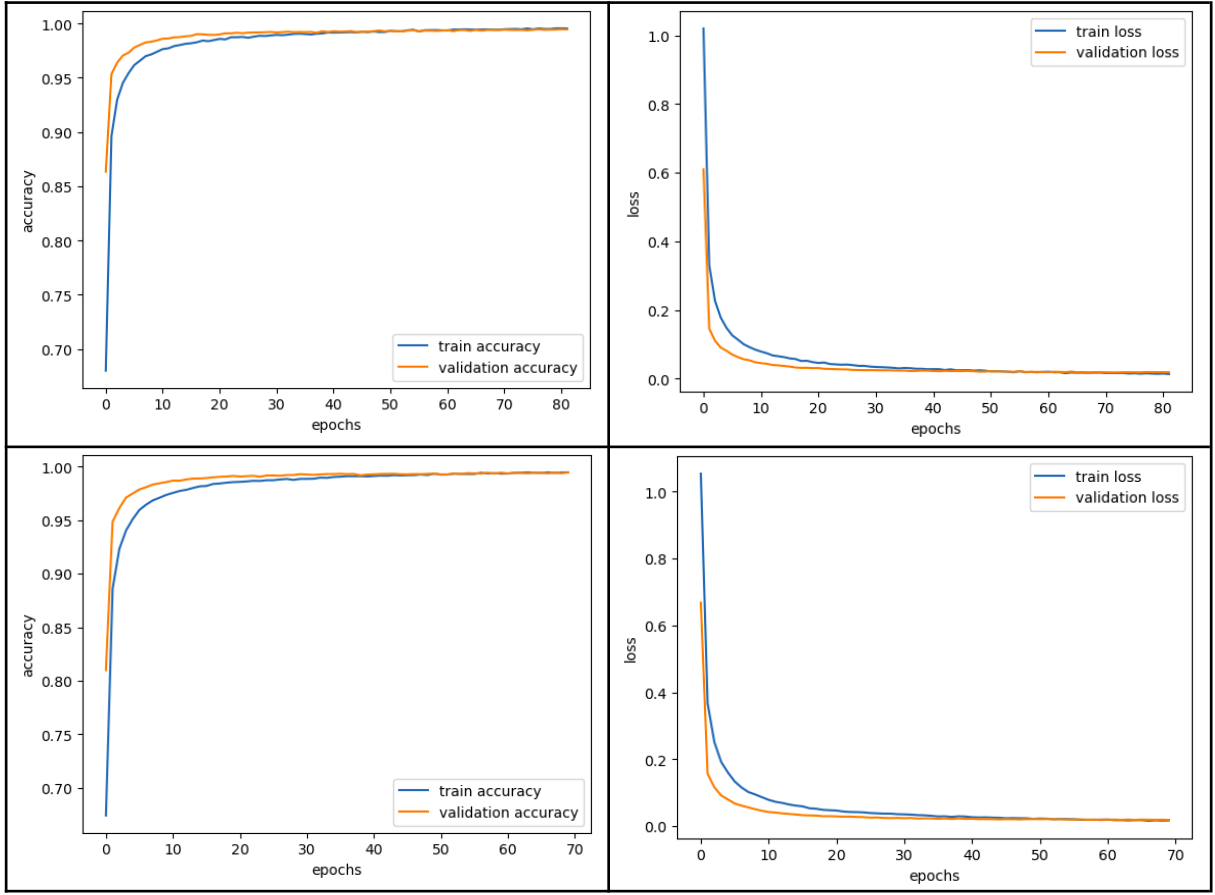

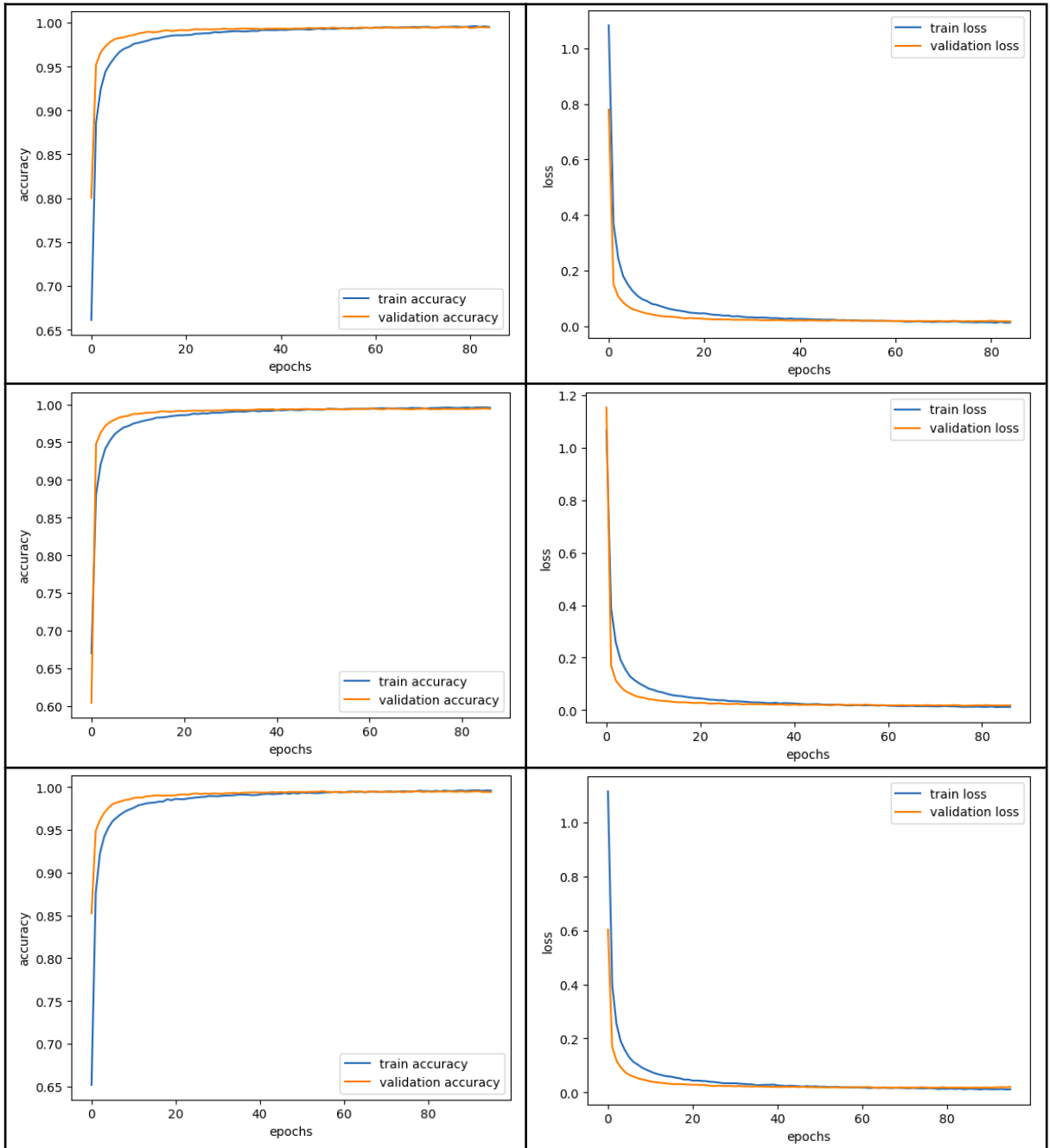

Figure 6: The training and validation accuracies and losses for the 5-fold cross-validation. The rows labeled 1 to 5 represent the five different folds used in the cross-validation.

The models generated from the five splits were evaluated on the testing set of 10,000 images. The models were evaluated on the training, validation, and testing datasets for all 5-fold. Figure 7 presents the confusion metrics for the five-fold models that were tested on the training, validation, and testing datasets. The metrics are provided in tabular form, and each row of the table represents a model that was tested on the dataset. Each column in the row represents the performance of the model on the corresponding dataset, i.e., training, validation, and testing.

The metrics provide a measure of the performance of the models in terms of their ability to correctly classify the input images into their respective classes.

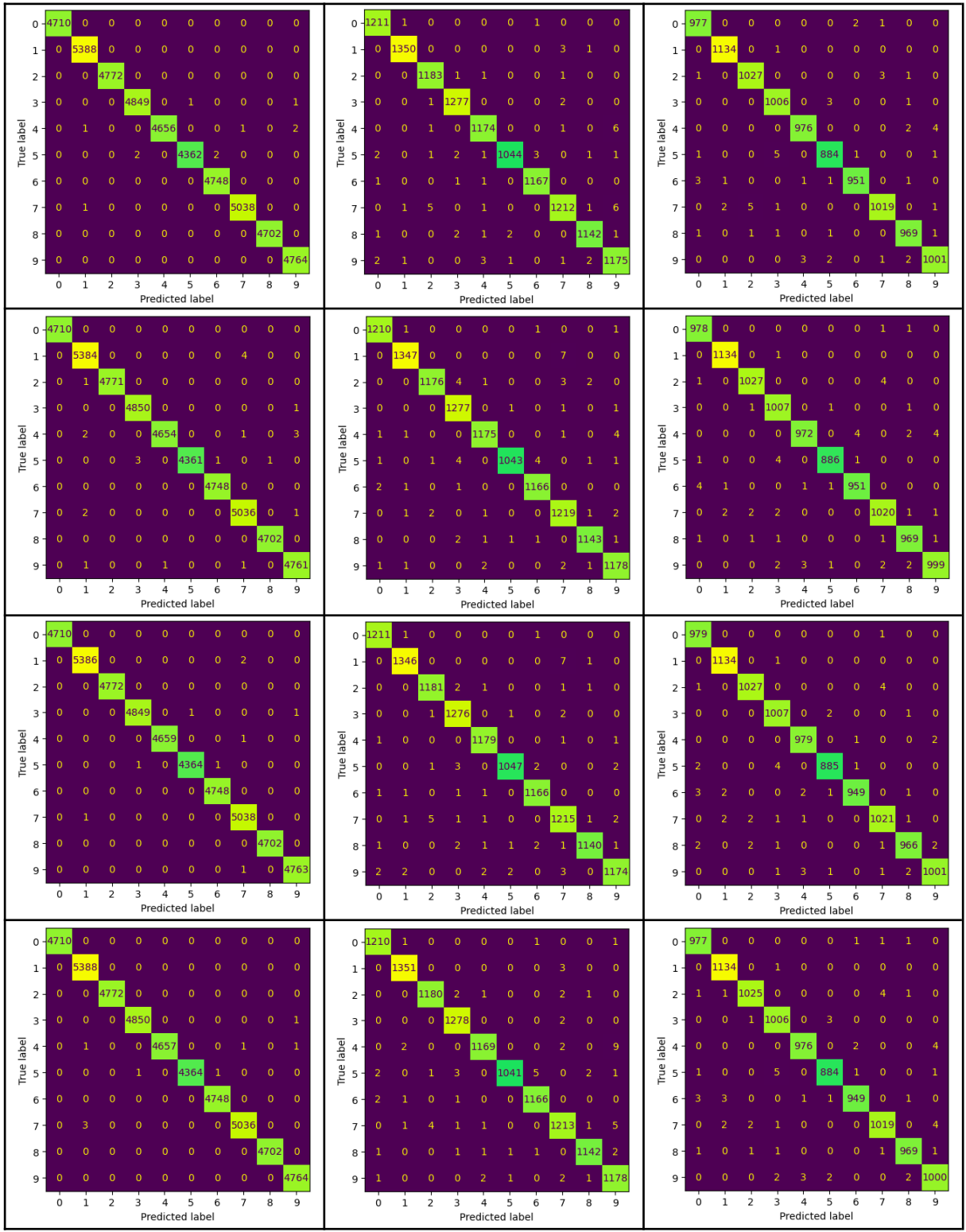

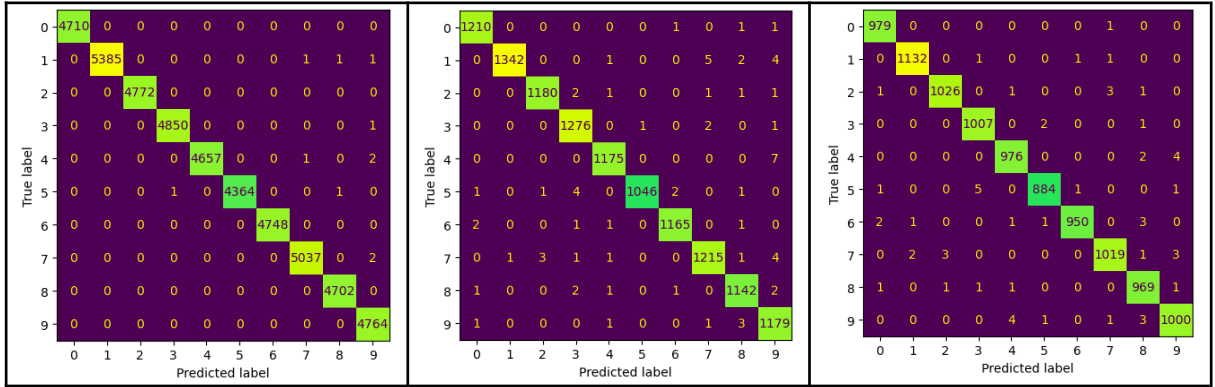

Figure 7: The confusion metrics for the five-fold models tested on the training, validation, and testing datasets. Each row represents a model, tested on the corresponding dataset in each column.

Additionally, the evaluation accuracies and losses for each of the five split models were provided in Table 3. The average accuracy for the test dataset was reported as 99.48, indicating that the model performed exceptionally well on the unseen test data. Well, the models were purposely evaluated on the training and evaluation datasets, and the results are tabulated in Table 4.

Table 3: The accuracies and losses of the models for each fold

| k-fold | Dataset | Accuracy | Loss |
|---|---|---|---|
| 1 | train | 0.9998 | 0.0014 |
| 1 | val | 0.9946 | 0.0183 |
| 1 | test | 0.9944 | 0.0161 |
| 2 | train | 0.9995 | 0.0022 |
| 2 | val | 0.9945 | 0.0184 |
| 2 | test | 0.9943 | 0.0166 |
| 3 | train | 0.9998 | 0.0011 |
| 3 | val | 0.9946 | 0.0174 |
| 3 | test | 0.9948 | 0.0169 |
| 4 | train | 0.9998 | 0.0010 |
| 4 | val | 0.9940 | 0.0185 |
| 4 | test | 0.9939 | 0.0191 |
| 5 | train | 0.9998 | 0.0012 |
| 5 | val | 0.9942 | 0.0208 |
| 5 | test | 0.9942 | 0.0192 |

Table 4: Average, Minimum, Maximum accuracies and losses of the model on the training, validation, and testing datasets.

| Datasets | 5-fold | | | | | |
|---|---|---|---|---|---|---|
| | Average Accuracy | Min Accuracy | Max Accuracy | Average Loss | Min Loss | Max Loss |
| train | 0.9997 | 0.9995 | 0.9998 | 0.00138 | 0.001 | 0.0022 |
| val | 0.9944 | 0.994 | 0.9946 | 0.01868 | 0.0174 | 0.0208 |
| test | 0.9943 | 0.9939 | 0.9948 | 0.01758 | 0.0161 | 0.0192 |

Our objective is to identify noisy images in the training dataset. To accomplish this, we have selected the 4th model from the 5-fold evaluation. This model exhibits low accuracy and high losses, making it suitable for our purpose. Upon analyzing this model, we discovered that a total of 81 images (72 from the validation set and 9 from the training set) were incorrectly classified out of 60,000 images. In addition, 61 images from the test data were also classified incorrectly. Refer to Tables 5 and 6 for the complete list of misclassified images from the training, validation, and testing datasets.

Table 5: Wrongly Classified images in Training Dataset of 60000 images

| Actual | Wrongly Classified images along with predictions and its confidence level (out of 1) |
|---|---|
| 0 | 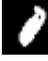 (1, 0.7), 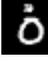 (6, 0.9), 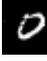 (9, 0.6) |
| 1 | 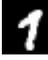 (7, 0.7), 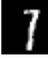 (7, 0.7), 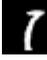 (7, 0.8) |
| 2 | 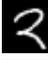 (3, 0.6), 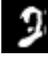 (3, 0.6), 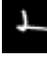 (4, 0.7), 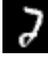 (7, 0.6), 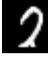 (7, 0.9), 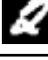 (8, 0.9) |
| 3 | 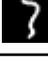 (7, 0.6), 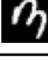 (7, 0.6), 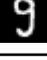 (9, 0.9) |
| 4 | 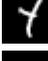 (1, 0.5), 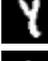 (1, 0.6), 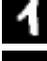 (1, 0.8), 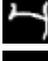 (7, 0.5), 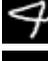 (7, 0.6), 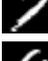 (7, 0.9), 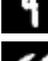 (9, 0.6), 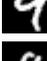 (9, 0.6), 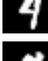 (9, 0.6), 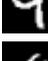 (9, 0.6), 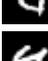 (9, 0.8), 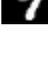 (9, 0.9), 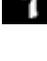 (9, 0.9), 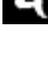 (7, 0.9), 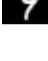 (9, 0.9), 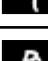 (9, 0.6) |
| 5 | 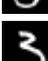 (0, 0.5), 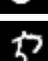 (0, 0.6), 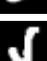 (2, 0.6), 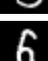 (3, 0.7), 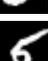 (3, 0.5), 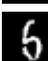 (3, 0.9), 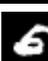 (3, 0.9), 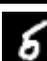 (6, 0.6), 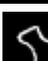 (6, 0.6), 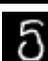 (6, 0.9), 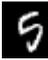 (6, 0.8), 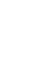 (6, 0.9), 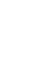 (6, 0.9), 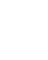 (8, 0.4), 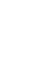 (8, 0.7), 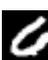 (9, 0.7) |
| 6 | 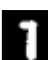 (0, 0.6), 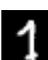 (0, 0.9), 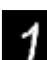 (1, 0.9), 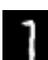 (3, 0.9) |
| 7 | 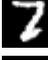 (1, 0.6), 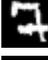 (1, 0.6), 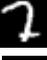 (1, 0.7), 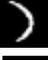 (1, 0.9), 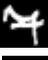 (2, 0.5), 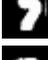 (2, 0.8), 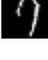 (4, 0.9), 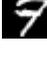 (8, 0.9), 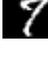 (3, 0.8), 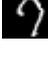 (4, 0.9), 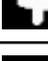 (8, 0.9), 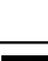 (9, 0.6), 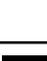 (9, 0.6), 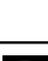 (9, 0.8), 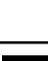 (9, 0.9), 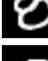 (9, 0.9) |
| 8 | 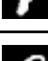 (0, 0.9), 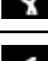 (3, 0.9), 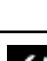 (4, 0.7), 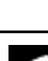 (5, 0.6), 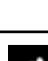 (6, 0.7), 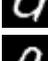 (9, 0.8), 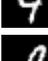 (9, 0.8) |
| 9 | 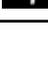 (0, 0.9), 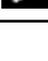 (4, 0.8), 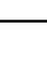 (4, 0.9), 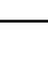 (5, 0.6), 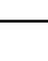 (7, 0.6), 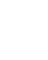 (7, 0.9), 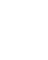 (8, 0.5) |

Table 6: Wrongly Classified images in Testing Dataset of 10000 images

| Actual | Wrongly Classified images - the confidence rate ranges between 0.3 and 0.99 |
|---|---|
| 0 | |
| 1 | |
| 2 | |
| 3 | |
| 4 | |
| 5 | |
| 6 | |
| 7 | |
| 8 | |
| 9 | |

As mentioned earlier, we observed that a total of 81 images (9 from the validation set and 72 from the training set) were misclassified out of 60,000 images. However, we also noted that there were some images that were correctly classified but with a confidence level lower or equal to 0.99. Although this threshold value was chosen arbitrarily, it implies that these correctly classified images may still contain noise or ambiguity.

For further analysis, please refer to Table 7, which presents the total count of noisy images in the training and validation datasets of all 5 fold evaluations.

Table 7: Noisy images in training dataset

| Actual | Fold 1 (Model 1) | | | Fold 2 (Model 2) | | | Fold 3 (Model 3) | | | Fold 4 (Model 4) | | | Fold 5 (Model 5) | | |
|---|---|---|---|---|---|---|---|---|---|---|---|---|---|---|---|
| | WC | CN | TC | WC | CN | TC | WC | CN | TC | WC | CN | TC | WC | CN | TC |
| 0 | 2 | 10 | 12 | 3 | 8 | 11 | 2 | 6 | 8 | 3 | 4 | 7 | 3 | 7 | 10 |
| 1 | 4 | 19 | 23 | 11 | 18 | 29 | 10 | 14 | 24 | 3 | 13 | 16 | 15 | 28 | 43 |
| 2 | 3 | 15 | 18 | 11 | 20 | 31 | 5 | 10 | 15 | 6 | 13 | 19 | 6 | 13 | 19 |
| 3 | 5 | 10 | 15 | 4 | 12 | 16 | 6 | 7 | 13 | 3 | 6 | 9 | 5 | 6 | 11 |
| 4 | 12 | 32 | 44 | 13 | 43 | 56 | 4 | 18 | 22 | 16 | 35 | 51 | 10 | 20 | 30 |
| 5 | 15 | 25 | 40 | 17 | 28 | 45 | 10 | 12 | 22 | 16 | 19 | 35 | 11 | 24 | 35 |
| 6 | 3 | 3 | 6 | 4 | 11 | 15 | 4 | 9 | 13 | 4 | 7 | 11 | 5 | 5 | 10 |
| 7 | 15 | 21 | 36 | 10 | 21 | 31 | 12 | 12 | 24 | 16 | 14 | 30 | 13 | 29 | 42 |
| 8 | 7 | 19 | 26 | 6 | 24 | 30 | 9 | 23 | 32 | 7 | 13 | 20 | 7 | 2 | 9 |
| 9 | 10 | 26 | 36 | 10 | 28 | 38 | 12 | 28 | 40 | 7 | 12 | 19 | 6 | 17 | 23 |
| Tot | 76 | 180 | 256 | 89 | 213 | 302 | 74 | 139 | 213 | 81 | 136 | 217 | 81 | 151 | 232 |

The following abbreviations are used in the table:
- WC: Incorrectly Classified Images
- CN: Correctly Classified Images with Noise, where the Confidence Level is less than 0.9
- TC: Total Count = WC + CN

Models 1, 2, 3, 4, and 5 have identified a total of 1286, 1651, 1216, 1182, and 1113 images as distorted respectively. There may be some overlap or unique identifications among the models. Consequently, the noisy images identified by all models will be removed from the training set, resulting in the removal of 489 images from the total of 60000 training and validation data. Table 8 displays the count of images in each class of the training and validation dataset before and after cleaning. The distribution of the cleaned training and validation dataset is illustrated in Figure 8.

Table 8: Dataset of Training and Validation

| Classes | Total images in Training and Validation dataset | Total images in cleaned Training and Validation dataset |
|---|---|---|
| 0 | 5923 | 5900 |
| 1 | 6742 | 6684 |
| 2 | 5958 | 5913 |
| 3 | 6131 | 6108 |
| 4 | 5842 | 5758 |
| 5 | 5421 | 5364 |
| 6 | 5918 | 5892 |
| 7 | 6265 | 6206 |
| 8 | 5851 | 5803 |
| 9 | 5949 | 5883 |
| **Total** | **60000** | **59511** |

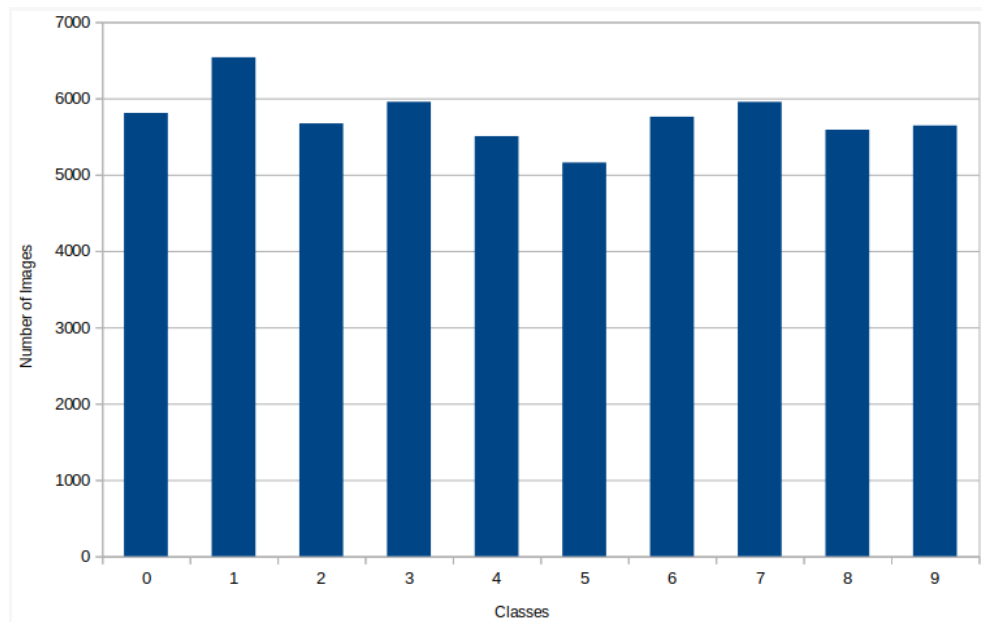

Figure 8: Distribution of digits in Cleaned Training Set

Despite the uneven distribution of data in the training and validation dataset, the model is still trained to evaluate the system's performance. The same steps as before are followed, including using a 5-fold evaluation. However, the data distribution in the split between the training and validation sets may differ in each of the 5-fold evaluations.

## 5.2  STAGE 2: TRAINING ON THE REDUCED TRAINING DATA

We removed 489 distorted images from the original training dataset of 60,000, resulting in a reduced dataset of 59,511 images. We trained the model using the same process as described in the previous section, with a training-validation split ratio of 80:20. This resulted in 47,609 images for training and 11,902 for validation. We also used 5-fold cross-validation in this case. Figure 9 displays the training and validation accuracies and losses for 5 iterations.

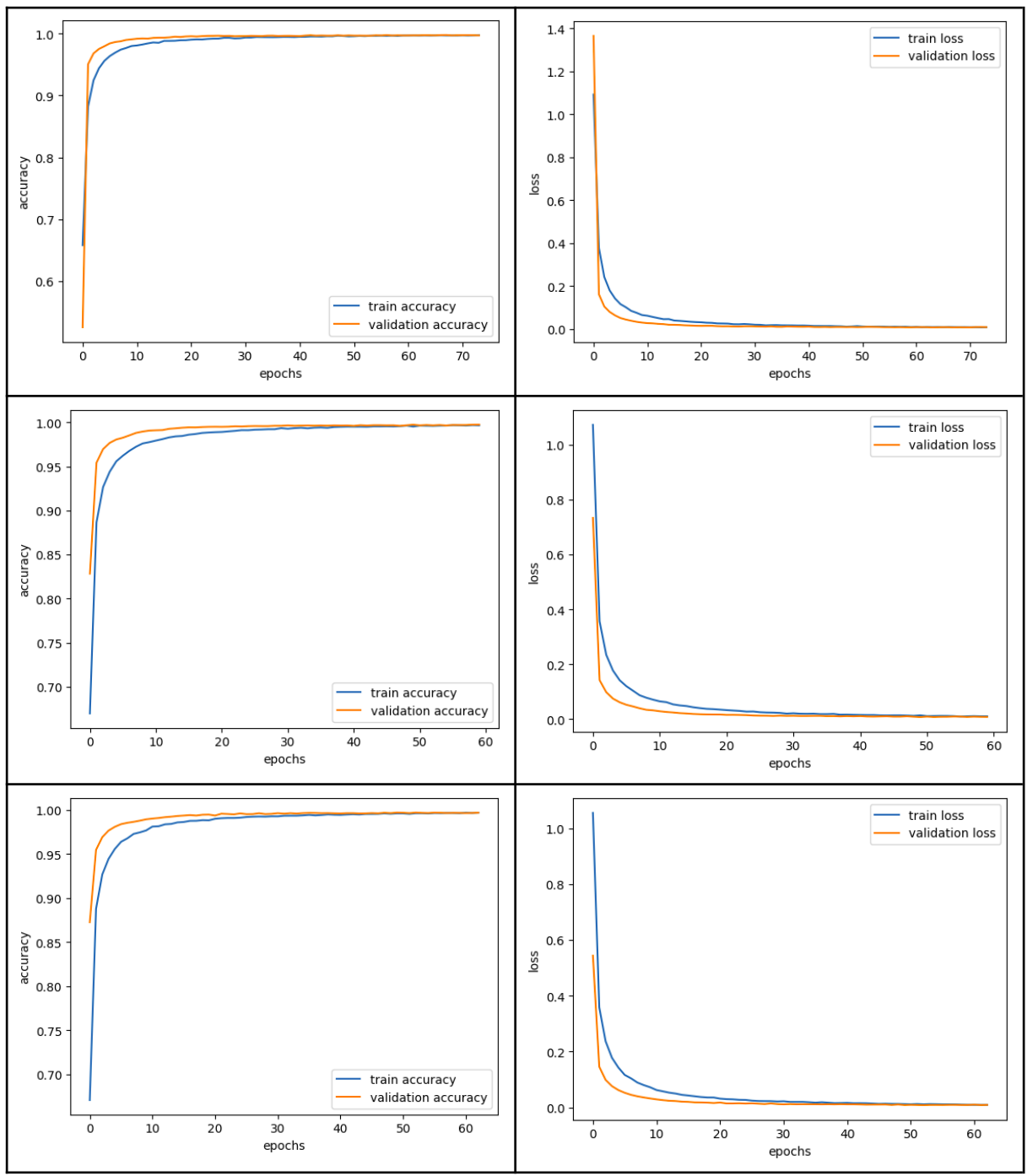

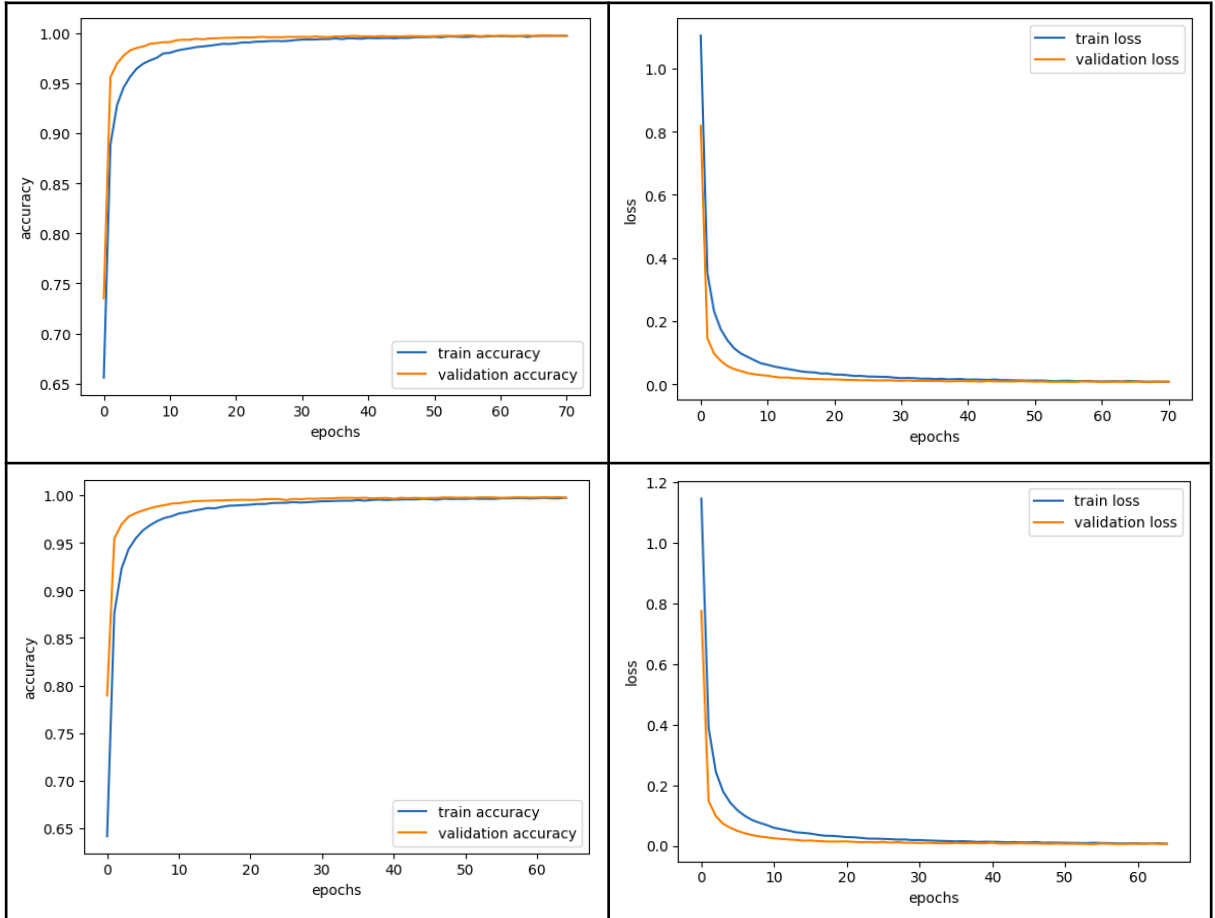

Figure 9: The training and validation accuracies and losses for the 5-fold cross-validation. The rows labeled 1 to 5 represent the five different folds used in the cross-validation.

After 40 epochs, the model reached convergence with a training accuracy of .999 and a validation accuracy of .9973, while the testing accuracy remained around .9943. Table 9 summarizes the losses and accuracies of the model across all 5-fold iterations for the training, validation, and testing datasets. Further, Table 10, shows the average accuracies and losses.

Table 9: The accuracies and losses of the models for each fold

| 5-fold | Dataset | Accuracy | Loss |
|---|---|---|---|
| 1 | train | 0.9999 | 0.0001 |
|   | val | 0.9971 | 0.0089 |
|   | test | 0.9935 | 0.0194 |
| 2 | train | 0.9999 | 0.0001 |
|   | val | 0.9975 | 0.0081 |
|   | test | 0.9945 | 0.0181 |
| 3 | train | 0.9999 | 0.0001 |
|   | val | 0.9969 | 0.0093 |
|   | test | 0.994 | 0.0194 |
| 4 | train | 0.9999 | 0.0001 |
|   | val | 0.9974 | 0.0085 |
|   | test | 0.9937 | 0.0194 |
| 5 | train | 0.9999 | 0.0001 |
|   | val | 0.9975 | 0.008 |
|   | test | 0.994 | 0.0183 |

Table 10: Average, Minimum, Maximum accuracies and losses of the model on the training, validation, and testing datasets.

| Datasets | 5-fold | | | | | |
| --- | --- | --- | --- | --- | --- | --- |
| | Average Accuracy | Min Accuracy | Max Accuracy | Average Loss | Min Loss | Max Loss |
| train | 0.9999 | 0.9999 | 0.9999 | 0.0001 | 0.0001 | 0.0001 |
| val | 0.9972 | 0.9969 | 0.9999 | 0.00856 | 0.008 | 0.0093 |
| test | 0.9939 | 0.9935 | 0.9945 | 0.01892 | 0.0181 | 0.0194 |

Figure 10 presents the confusion metrics for the five-fold models that were tested on the training, validation, and testing datasets. The metrics are provided in tabular form, and each row of the table represents a model that was tested on the dataset. Each column in the row represents the performance of the model on the corresponding dataset, i.e., training, validation, and testing.

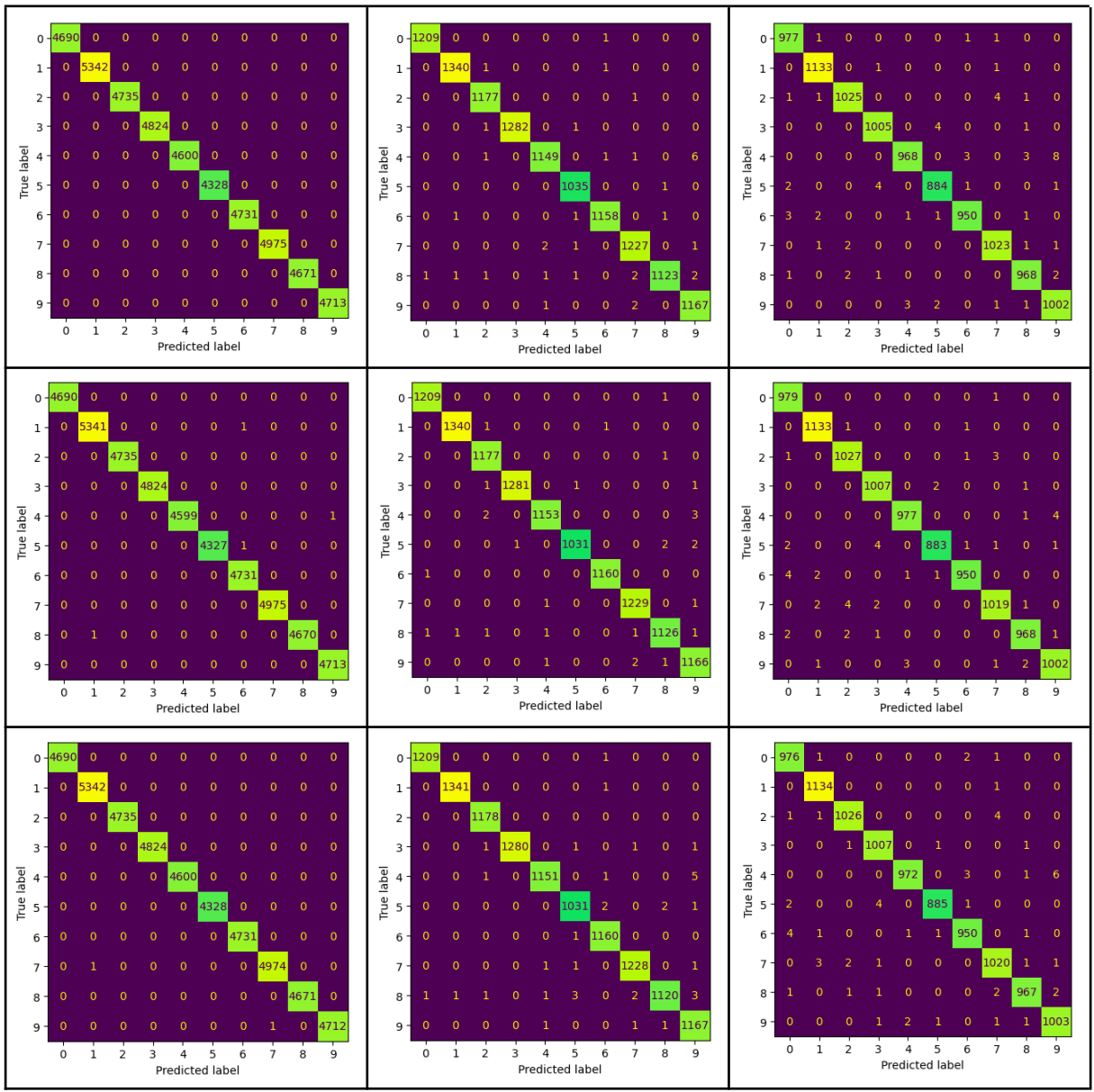

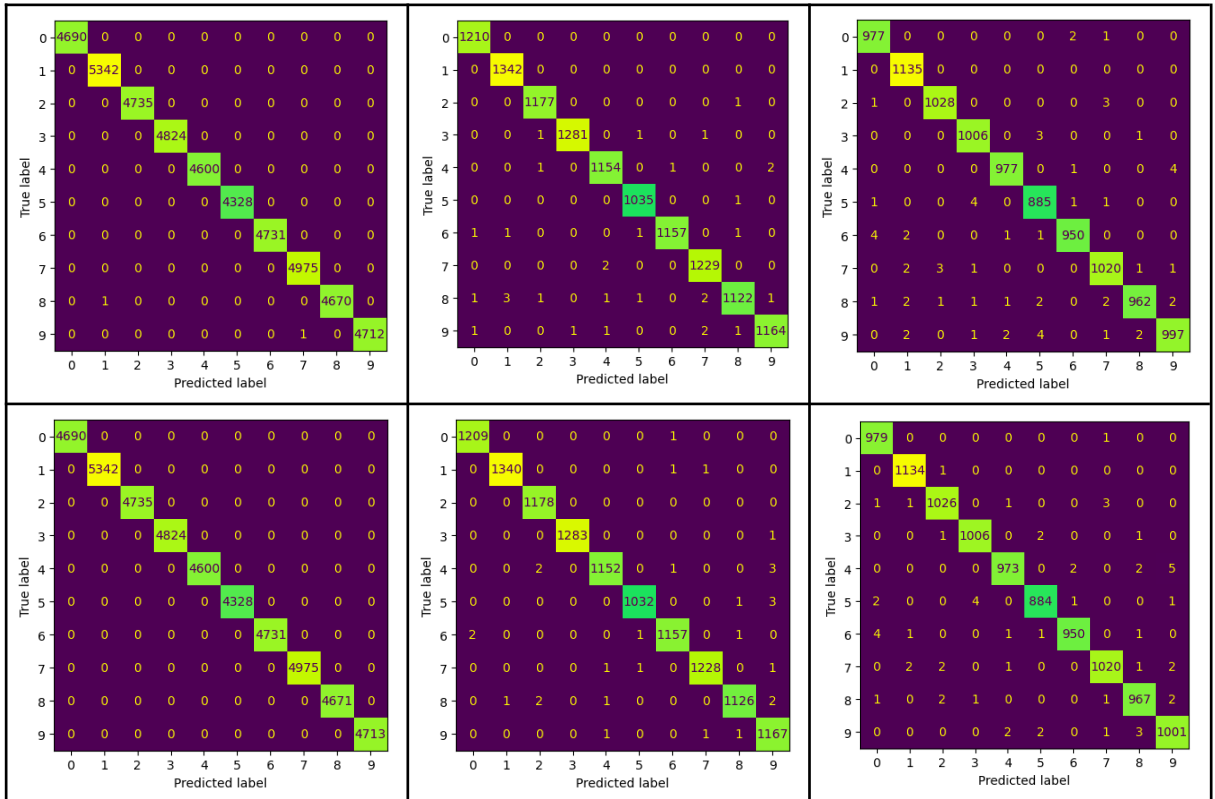

Figure 10: The confusion metrics for the five-fold models tested on the training, validation, and testing datasets. Each row represents a model, tested on the corresponding dataset in each column.

Tables 11 and 12 are for the complete list of misclassified images from the training, validation, and testing datasets.

Table 11: Wrongly Classified images in Training Dataset of 59511 images

| Actual | Wrongly Classified images |
|---|---|
| 0 | |
| 1 | |
| 2 | |
| 3 | |
| 4 | |
| 5 | |
| 6 | |
| 7 | |
| 8 | |
| 9 | |

Table 12: Wrongly Classified images in Testing Dataset of 10000 images

| Actual | Wrongly Classified images - the confidence rate ranges between 0.3 and 0.99 |
|---|---|
| 0 | 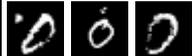 |
| 1 | 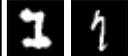 |
| 2 | 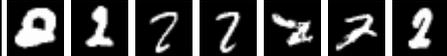 |
| 3 | 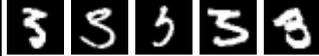 |
| 4 | 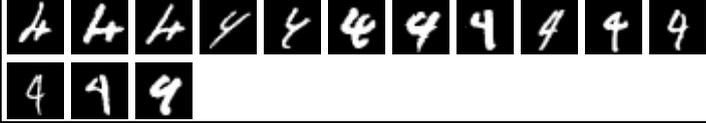 |
| 5 | 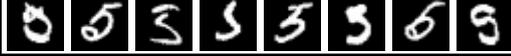 |
| 6 | 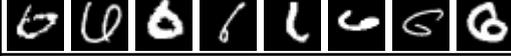 |
| 7 | 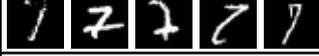 |
| 8 | 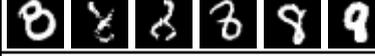 |
| 9 | 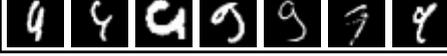 |

As previously mentioned, a total of 81 images (9 from the validation set and 72 from the training set) were misclassified out of 60,000 images.

Our results demonstrate that removing the distorted images improved training and validation accuracy, both above .999 and .997, respectively. We also observed misclassifications in classes 4 and 9, which may be attributed to the edges and curves in those digits.

### 5.3 FAILURE CASE - A CASE STUDY

The proposed model has one limitation: it is a rotation variant, meaning that the model cannot effectively process rotated digits. Table 13 displays some examples of such images.

Table 13: Wrongly Classified images in Testing Dataset of 10000 images because of rotation

| Actual | Image | Predicted |
|---|---|---|
| 8 | 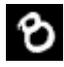 | 0 |
| 4 | 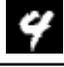 | 9 |
| 6 | 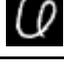 | 0 |

### 5.4 COMPARATIVE ANALYSIS

Table 14 presents a comparative analysis of recent papers on the topic. Our proposed model provides the following distinct advantages when compared to these papers.
1. First, no-augmented data for building the deep network for achieving maximum accuracy.

2. Second, the predicted classes' confidence level has been improved without or less distorted or noisy, or ambiguous images. Most of the other papers either do not discuss or provide limited information on the confidence level of each predicted image.
3. Additionally, while many of the proposed models utilized the entire MNIST training dataset of 60,000 images for training, our model used only 80% of the training data for actual training and reserved the remaining 20% for validation.

Although some of the literature shows an accuracy above 99.5% on the testing dataset, our discussions indicate that these high accuracies can make it difficult for humans to distinguish between the images. This suggests that many of these papers may have trained their models on the entire training data without filtering out the noise images, which can lead to overfitting.

Table 14: Comparative analysis - State of the art

| Approach | Testing Dataset |
| --- | --- |
| Dynamic Routing Between Capsules, 2017 [11] | 99.75% |
| Lets keep it simple, Using simple architectures to outperform deeper and more complex architectures, 2016 [54] | 99.75% |
| Batch-Normalized Maxout Network in Network, 2015 [35] | 99.76% |
| APAC:Augmented Pattern Classification with Neural Networks, 2015 [28] | 99.77% |
| Multi-Column Deep Neural Networks for Image Classification, 2012 [27] | 99.77% |
| No routing needed between capsules, 2021 [19] | 99.83% |
| Ensembles: Regularization of Neural Networks using DropConnect 2013 [26] | 99.79% |
| Ensembles: RMDL:Random Multimodel Deep Learning for Classification 2018 [29] | 99.82% |
| Ensembles:No routing needed between capsules, 2021 [19] | 99.87% |
| **Ours** | **99.75 (Cleaned Validation) & 99.43 (Testing)** |

5.5 CONTRADICTION STUDY

In their publication from 2021 [19], the authors emphasized the presence of ambiguous and distorted images, which they successfully addressed to achieve the highest accuracy compared to other existing literature. Similarly, the original dataset paper [2] also acknowledged the existence of such distorted images, which inherently complicates the recognition task. While the paper from 2021 [19] outperformed other state-of-the-art models in terms of accuracy, it is clear that the trained models exhibit bias and lack generalization capabilities.

6 CONCLUSION

In summary, this study demonstrates that removing distorted images can lead to a significant increase in classification accuracy and confidence level. Our novel deep network model, applied to the MNIST dataset, improved the validation accuracy from 99.44% to 99.72%. The model also effectively addressed overfitting and underfitting issues, as evidenced by the increased confidence level for correctly classified images and decreased confidence level for misclassified ones. Although our research was limited to the MNIST dataset, it can be extended to other datasets. Additionally, we used an 8:2 ratio for training and validation, as suggested by previous literature, but this ratio can be varied in future research. To achieve even higher accuracy on the testing data, future work may involve further reducing the training data and exploring data augmentation techniques for image rotations. While some literature may demonstrate higher accuracy rates than our proposed approach, it is evident that our model overcomes bias and exhibits generalized capability. This paper emphasizes the importance of data quality and preprocessing in developing accurate deep-learning models for image classification tasks, and our proposed approach shows promise in improving confidence levels in model predictions.